\documentclass[11pt,letterpaper]{article}
\usepackage[letterpaper]{geometry}
\usepackage{acl2012}
\usepackage{times}
\usepackage{latexsym}
\usepackage{amsmath}
\usepackage{amssymb}
\usepackage{graphicx}
\usepackage{algorithm2e}
\usepackage{algpseudocode}
\usepackage{ifthen}
\usepackage[utf8]{inputenc}
\usepackage{url}
\usepackage{booktabs}

    \usepackage{amsmath,amssymb,amsfonts,amsthm, todonotes}

    \newcommand{\EE}{\mathbb{E}}

    \usepackage{array,multirow}
    \newcolumntype{C}[1]{>{\centering\arraybackslash}m{#1}}

   \let\<\textless
   \let\>\textgreater
   \usepackage[T1]{fontenc}

   \makeatletter
\newcommand{\@BIBLABEL}{\@emptybiblabel}
\newcommand{\@emptybiblabel}[1]{}
\makeatother
\usepackage[hidelinks]{hyperref}

\newcommand\sN{\ensuremath{\mathcal{N}}}

\newcommand\sX{\ensuremath{\mathcal{X}}}

      \newcommand\eqdef{\ensuremath{\stackrel{\rm def}{=}}}

\newcommand\refsec[1]{Section~\ref{sec:#1}}

\newcommand\reftab[1]{Table~\ref{tab:#1}}

\ifthenelse{\isundefined{\definition}}{\newtheorem{definition}{Definition}}{}
\ifthenelse{\isundefined{\assumption}}{\newtheorem{assumption}{Assumption}}{}
\ifthenelse{\isundefined{\hypothesis}}{\newtheorem{hypothesis}{Hypothesis}}{}
\ifthenelse{\isundefined{\proposition}}{\newtheorem{proposition}{Proposition}}{}
\ifthenelse{\isundefined{\theorem}}{\newtheorem{theorem}{Theorem}}{}
\ifthenelse{\isundefined{\lemma}}{\newtheorem{lemma}{Lemma}}{}
\ifthenelse{\isundefined{\corollary}}{\newtheorem{corollary}{Corollary}}{}
\ifthenelse{\isundefined{\alg}}{\newtheorem{alg}{Algorithm}}{}
\ifthenelse{\isundefined{\example}}{\newtheorem{example}{Example}}{}

\newcommand\sLLex{\text{LEX}}

\title{Generating Sentences by Editing Prototypes}

 \author{Kelvin Guu*$^2$ ~~ Tatsunori B. Hashimoto*$^{1,2}$ ~~ Yonatan Oren$^1$ ~~ Percy Liang$^{1,2}$ \\
   \textbf{(* equal contribution)} \\
   $^1$Department of Computer Science ~~ $^2$Department of Statistics \\
   Stanford University \\
   {\small \tt \{kguu,thashim,yonatano\}@stanford.edu ~~ pliang@cs.stanford.edu} \\
 }

\date{}

\begin{document}

\maketitle

\begin{abstract}
We propose a new generative language model for sentences that first samples a prototype sentence from
the training corpus and then edits it into a new sentence.
Compared to traditional language models that generate from scratch either left-to-right
or by first sampling a latent sentence vector,
our prototype-then-edit model improves perplexity
on language modeling and generates higher quality outputs according to human
evaluation. Furthermore, the model gives rise to a latent edit vector that
captures interpretable semantics such as sentence similarity and sentence-level analogies.
 \end{abstract}

\section{Introduction}

The ability to generate sentences is core to many NLP tasks, including
machine translation, summarization, speech recognition, and dialogue.
Most neural models for these tasks are based on
recurrent neural language models (NLMs), which generate sentences from scratch,
often in a left-to-right manner \cite{bengio2003neural}.
It is often observed that such NLMs suffer from the problem of
favoring generic utterances such as ``I don't know'' \cite{li2016diversity}.
At the same time, naive strategies to increase diversity have been shown to
compromise grammaticality \cite{shao2017generating}, suggesting that current NLMs may
lack the inductive bias to faithfully represent the full diversity of complex utterances.

\begin{figure}
  \centering
  \includegraphics[scale=0.37]{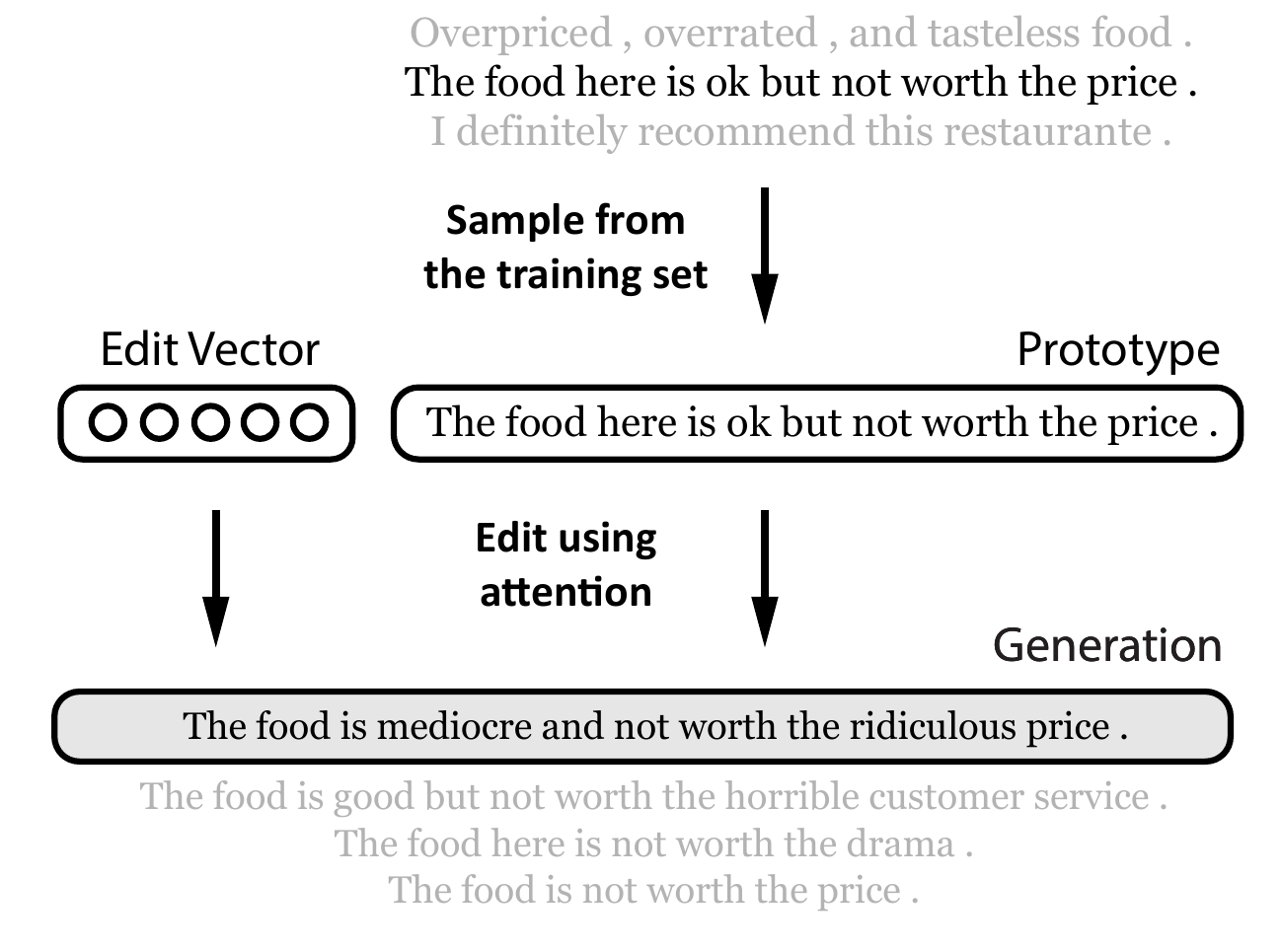}
  \vspace{-25pt}
\caption{The prototype-then-edit model generates a sentence by sampling a
  random example from the training set and then editing it using a randomly sampled edit vector.
}

\label{fig:one}
  \vspace{-10pt}
\end{figure}

Indeed, it is difficult even for humans to write complex text
from scratch in a single pass; we often create an initial draft
and incrementally revise it \cite{hayes1986writing}.
Inspired by this process, we propose a new unconditional generative model of text which we
call the prototype-then-edit model, illustrated in Figure~\ref{fig:one}.
It first samples a random \emph{prototype} sentence from the training corpus,
and then invokes a \emph{neural editor},
which draws a random ``edit vector'' and generates a new sentence by attending to the prototype while conditioning on the edit vector.
The motivation is that sentences from the corpus provide a high quality starting point: they are
grammatical, naturally diverse, and exhibit no bias towards shortness or
vagueness. The attention mechanism \cite{bahdanau2015neural} of the neural editor
strongly biases the generation towards the prototype, and therefore it needs to
solve a much easier problem than generating from scratch.

We train the neural editor by maximizing an approximation to the
generative model's log-likelihood.  This objective is a sum over
lexically-similar sentence pairs in the training set, which we
can scalably approximate using locality sensitive hashing.
We also show empirically that most lexically similar sentences are also
semantically similar, thereby endowing the neural editor with additional semantic
structure.  For example, we can use the neural editor to perform a random walk
from a seed sentence to traverse semantic space.

We compare our prototype-then-edit model
to approaches that generate from scratch on both
language generation quality and semantic properties.
For the former, our model generates higher quality generations according to human evaluations,
and improves perplexity by 13 points on the Yelp corpus and 7 points on the
One Billion Word Benchmark.
For the latter, we show that latent edit vectors outperform
standard sentence variational autoencoders \cite{bowman2016continuous}
on semantic similarity, locally-controlled text generation, and a sentence analogy task.
 \section{Problem statement}
\label{sec:problem}

Our primary goal is to learn a generative model of sentences for use as a language model.\footnote{
For many applications such as machine translation or dialogue generation,
there is a context (e.g. foreign sentence, dialogue history),
which can be supplied to both the prototype selector and the neural editor.
This paper focuses on the unconditional case, proposing an alternative to LSTM based language models.
}
In particular, we model sentence generation as a prototype-then-edit process:

\begin{enumerate}
\itemsep0em	
\item \textbf{Select prototype}: Given a training corpus of sentences $\mathcal{X}$, randomly sample a
  \emph{prototype} sentence $x'$ from a \emph{prototype distribution} $p(x^\prime)$
  (in our case, uniform over $\mathcal{X}$).
\item \textbf{Edit}: Sample an edit vector $z$ (encoding the type of edit to be performed)
  from an \emph{edit prior} $p(z)$.
    Then, feed the edit vector $z$ and the previously selected prototype $x'$ into a
    \emph{neural editor} $p_\text{edit}(x \mid x^{\prime}, z)$,
   which generates a new sentence $x$.
    \end{enumerate}

Under this model, the likelihood of a sentence is:
\begin{align}
  p(x) &= \sum_{x^{\prime}\in\mathcal{X}} p(x \mid x^\prime) p(x^\prime) \label{eq:dist} \\
  p(x \mid x^\prime) &= \EE_{z \sim p(z)}\left[ p_\text{edit}(x\mid x^{\prime},z) \right],  \label{eq:cond}
\end{align}
where both prototype $x'$ and edit vector $z$ are latent variables.

Our formulation stems from the observation that many sentences in a large corpus
can be represented as minor transformations of other sentences. For
example, in the Yelp restaurant review corpus \cite{yelp2017yelp} we find that 70\% of the test set is within word-token Jaccard distance 0.5 of a training set sentence, even
though almost no sentences are repeated verbatim. This implies that a neural editor which models lexically similar sentences should be an effective generative model for large parts of the test set.

A secondary goal for the neural editor is to capture certain semantic properties;
we focus on the following two in particular:
\begin{enumerate}
\itemsep0em
\item Semantic smoothness: an edit should be able to alter the semantics
of a sentence by a small and well-controlled amount, while multiple edits should
make it possible to accumulate a larger change.
\item Consistent edit behavior: the edit vector $z$ should
model/control the variation in the type of edit that is performed.
When we apply the same edit vector on different sentences,
the neural editor should perform \emph{semantically analogous} edits across the sentences.
\end{enumerate}

In \refsec{experiments}, we show that the
neural editor can successfully capture both properties, as reported by human evaluations.
 \section{Approach}

\newcommand{\pedit}{p_\text{edit}(x\mid x',z)}
\newcommand{\qedit}{q(z \mid x', x)}
\newcommand{\vmf}{\text{vMF}}
\newcommand{\kl}{\text{KL}}
\newcommand{\elbo}{\text{ELBO}(x,x')}
\newcommand{\xp}{x^{\prime}}

We would like to train our neural editor $\pedit$ by maximizing the marginal likelihood
(Equation \ref{eq:dist}) via gradient ascent, but the objective cannot be computed exactly because it
involves a sum over all prototypes $x'$ (expensive) and an expectation over
the edit prior $p(z)$ (no closed form).

We therefore propose two approximations to overcome these challenges:
\begin{enumerate}
\itemsep0em
\item We lower bound the sum over latent prototypes $x'$ (in Equation \ref{eq:dist}) by only summing
over $x'$ that are \emph{lexically similar} to $x$.
\item We lower bound the expectation over the edit prior (in Equation \ref{eq:cond}) using
  the evidence lower bound (ELBO) \cite{jordan1999variational,doersch2016tutorial} which can be effectively approximated.
\end{enumerate}
We describe and motivate these approximations in Sections \ref{sec:lowerbound} and \ref{sec:learn_to_edit}, respectively.
In Section \ref{sec:combine}, we combine the two approximations to give the final objective.
Sections \ref{sec:vae-arch} and \ref{sec:q-discuss} drill down further into our specific model architecture.

\subsection{Approximate sum on prototypes, $x^\prime$}
\label{sec:lowerbound}

Equation \ref{eq:dist} defines the probability of generating a sentence $x$ as the total
probability of reaching $x$ via edits from every prototype $x' \in \mathcal{X}$. However, most
prototypes are unrelated and should have very small probability of transforming into $x$.
Therefore, we approximate the summation over prototypes by only
considering prototypes $x'$ that have
\emph{high lexical overlap} with $x$.
To that end, define a lexical similarity neighborhood as:
\begin{align*}
\mathcal{N}(x) &\eqdef \{x'\in\mathcal{X}:d_J(x,x') < 0.5\},
\end{align*}
where $d_J(x, x')$ is the Jaccard distance between $x$ and $x'$ (treating each as a set of word tokens).

We will now lower bound $\log p(x)$ in two ways:
(i) we will sum over only prototypes in the neighborhood $\sN(x)$ rather than
over the entire training set $\sX$ as discussed above;
(ii) we will push the log inside the summation using Jensen's inequality, as is standard with variational lower bounds.
Recall that the distribution over prototypes is uniform ($p(x') = 1/|\sX|$),
and define $R(x) = \log(|\sN(x)|/|\sX|)$.
The derivation is as follows:
\resizebox{0.47\textwidth}{!}{
\begin{minipage}{1.2\linewidth}
\begin{align}
\log p(x)
  &= \log\left[\sum_{x^{\prime}\in\mathcal{X}}p(x\mid x^{\prime}) p(x^{\prime})\right] \nonumber \\
  &\stackrel{(i)}{\geq} \log\left[\sum_{x^{\prime}\in\mathcal{N}(x)}p(x\mid x^{\prime}) p(x^{\prime})\right] \label{eq:lowerbound} \\
  &= \log\left[ |\sN(x)|^{-1} \sum_{x^{\prime}\in\mathcal{N}(x)} p(x\mid x^{\prime}) \right] + R(x) \nonumber \\
  &\stackrel{(ii)}{\ge} |\sN(x)|^{-1} \underbrace{\sum_{x^{\prime}\in\mathcal{N}(x)} \log p(x\mid x^{\prime})}_{\eqdef \sLLex(x)} + R(x). \nonumber 
\end{align}
\end{minipage}}
Assuming the neighborhood size $|\sN(x)|$ is constant across $x$,
then $\sLLex(x)$ is a lower bound of $\log p(x)$ up to constants.
For each $x$, the neighborhood $\mathcal{N}(x)$ can be efficiently precomputed with
locality sensitive hashing (LSH) and minhashing. The full procedure is
described in Appendix \ref{sec:lsh}.

Note that $\sLLex(x)$ is still intractable to compute because each $\log p(x|x')$ term
involves an expectation over the edit prior $p(z)$ (Equation \ref{eq:cond}).
We address this in Section \ref{sec:learn_to_edit},
but first, an interlude.

\paragraph{Interlude:~lexical similarity semantics.}

So far,
we have motivated lexical similarity neighborhoods via computational considerations,
but we found that lexical similarity training also captures semantic similarity.
One can certainly construct sentences with small lexical distance that
differ semantically (e.g., insertion of the word ``not'').
However, since we mine sentences from a corpus grounded in real world events,
most lexically similar sentences are also semantically similar.
For example, given ``my son enjoyed the delicious pizza'', we are far more
likely to see ``my son enjoyed the delicious macaroni'', versus ``my son hated the delicious pizza''. 

Human evaluations of 250 edit pairs sampled from lexical similarity neighborhoods on the Yelp corpus
support this 
conclusion. 35.2\% of the sentence pairs were judged to be exact paraphrases,
while 84\% of the pairs were judged to be at least roughly equivalent. Sentence
pairs were negated or change in topic only 7.2\% of the time.  Thus, a neural editor
trained on this distribution should preferentially generate semantically similar edits.

Note that semantic similarity is not needed if we are only interested in modeling the distribution $p(x)$.
But it does enable us to learn an edit model $p(x | x')$ that prefers semantically meaningful edits,
which we explore in Section \ref{sec:edit-results}.

\subsection{Approximate expectation on edit vectors, $z$}
\label{sec:learn_to_edit}

In \refsec{lowerbound}, we approximated the marginal likelihood $\log p(x)$ by $\sLLex(x)$,
which is a summation over terms of the form:
\begin{align}
\label{eq:log_edit}
\log p(x \mid x') = \log \EE_{z \sim p(z)}\left[ \pedit \right].
\end{align}
Unfortunately the expectation over $p(z)$ has no closed form, and naively approximating it by Monte Carlo sampling $z \sim p(z)$
will have unacceptably high variance, because $\pedit$ will be almost zero for nearly all $z$ sampled
from $p(z)$, while being large for a few important but rare values.

To address this, we introduce an \emph{inverse neural editor} $\qedit$:
given a prototype $x'$ and a revised sentence $x$, it generates edit vectors
that are \emph{likely} to map $x'$ to $x$, concentrating probability on the few rare but important values of $z$.

We can then use the evidence lower bound (ELBO) to lower bound Equation \ref{eq:log_edit}:
\newcommand{\Lgen}{\mathcal{L}_\text{gen}}
\newcommand{\LKL}{\mathcal{L}_\text{KL}}
\begin{align*}
\log p(x | x') &\geq \underbrace{\EE_{z\sim q(z\mid x',x)}\left[\log \pedit \right]}_{\Lgen} \\
&\quad - \underbrace{\kl(q(z\mid x',x)\ \| \ p(z))}_{\LKL} \\
&\eqdef \elbo.
\end{align*}
Since $\Lgen$ is an expectation over $\qedit$ instead of $p(x)$, it
can be effectively Monte Carlo estimated by sampling $z \sim \qedit$.
The second term, $\LKL$, penalizes the difference between $\qedit$ and $p(x)$,
which is necessary for the lower bound to hold.
A thorough introduction to the ELBO is provided in \newcite{doersch2016tutorial}.

Note that $\qedit$ and $\pedit$ combine to form a \emph{variational autoencoder} (VAE) \cite{kingma2014variational},
where $\qedit$ is the \emph{variational encoder} and $\pedit$ is the \emph{variational decoder}.

\subsection{Final objective}
\label{sec:combine}

Combining the lower bounds $\sLLex(x)$ and $\elbo$, our final approximation of the log-likelihood is
\begin{align*}
\sum_{x'\in \mathcal{N}(x)}\elbo.
\end{align*}

We optimize this objective using stochastic gradient ascent with respect to $\Theta = (\Theta_p, \Theta_q)$,
where $\Theta_p$ are the parameters for the neural editor and $\Theta_q$ are the parameters
for the inverse neural editor.

\subsection{Model architecture}
\label{sec:vae-arch}

To recap, our model features three components:
the \textbf{neural editor} $\pedit$,
the \textbf{edit prior} $p(z)$,
and the \textbf{inverse neural editor} $q(z \mid x',x)$.
We detail each of these components below.

\paragraph{Neural editor $\pedit$.}
We implement our neural editor as a left-to-right sequence-to-sequence model
with attention, where the prototype $x'$ is the input sequence and the revised
sentence $x$ is the output sequence. We employ an encoder-decoder architecture similar to Wu \shortcite{wu2016google},
extending it to condition on an edit vector $z$ by concatenating $z$ to the input
of the decoder at each time step.

The prototype encoder is a 3-layer bidirectional LSTM. The inputs to each
layer are the concatenation of the forward and backward hidden states of the previous layer,
with the exception of the first layer, which takes word vectors initialized using GloVe \cite{pennington2014glove}.

The decoder is a 3-layer LSTM with attention. At each
time step, the hidden state of the top layer is used to compute attention 
over the top-layer hidden states of the prototype encoder. The resulting attention context vector
is then concatenated with the decoder's top-layer hidden state and used to compute
a softmax distribution over output tokens.

\paragraph{Edit prior $p(z)$.} We sample the edit vector $z$ from the prior by first sampling its
scalar length $z_\text{norm} \sim \text{Unif}(0, 10)$ and then sampling its direction
$z_\text{dir}$ (a unit vector) from the uniform distribution on the unit sphere. The resulting
$z = z_\text{norm} \cdot z_\text{dir}$. As we will see later, this particular choice
of the prior enables us to easily compute $\LKL$.

\paragraph{Inverse neural editor $\qedit$.}

Given an edit pair $(x', x)$, the inverse neural editor
must infer what vectors $z$ are likely to map $x'$ to $x$.

Suppose that $x'$ and $x$ only differed by a single word $w$.
Then one might propose that the edit vector $z$ should be equal
to the word vector for $w$.
Generalizing this intuition to multi-word edits, 
we would like multi-word insertions to be
represented as the sum of the inserted word vectors, and similarly for deletions.

Formally, define
$I = x \backslash x'$ to be the set of words added to $x'$, and $D = x' \backslash x$ to be the words deleted.
We represent the difference between $x'$ and $x$ using the following vector:
\[f\left(x,x^{\prime}\right)=\sum_{w\in I}\Phi\left(w\right)\oplus\sum_{w\in D}\Phi\left(w\right)\]
where $\Phi(w)$ is the word vector for word $w$ and $\oplus$ denotes
concatenation. The word embeddings $\Phi$ are parameters of $q$. In our work,
we initialize $\Phi(w)$ to be 300-dimensional GloVe vectors.

\begin{figure}[h!]
\centering
\includegraphics[scale=0.24,trim={0 70 0 0},clip]{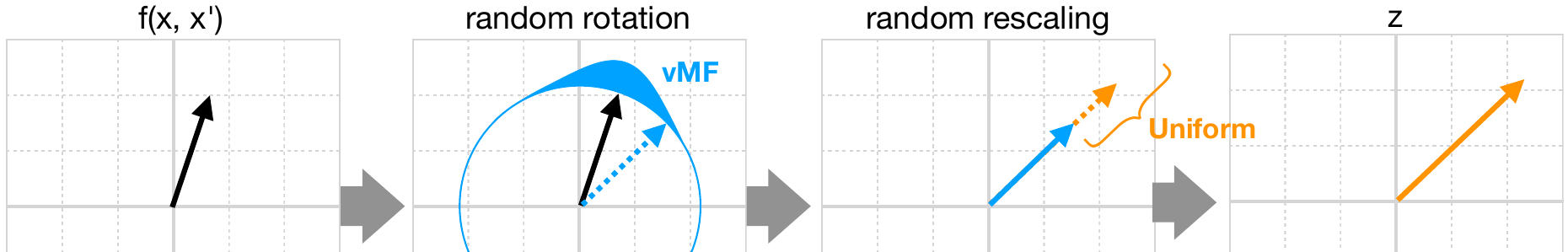}
\caption{
The inverse neural editor $q$ outputs a perturbed version of $f(x, x')$. The perturbation
process is a random rotation (according to the von-Mises Fisher distribution) followed by a random rescaling
(according to the uniform distribution).
}
\label{fig:vmf}
\end{figure}

Since we construct our edit vectors as the sum of word vectors, and similarities between word vectors have
traditionally been measured with cosine similarity, we design $q$ to add noise to perturb the direction of the vector $f$.
In particular, a sample from $q$ is simply a perturbed version of $f$:
obtained by adding von-Mises Fisher (vMF) noise,
and we perturb the magnitude of $f$ by adding uniform noise.
We visualize this perturbation process in Figure \ref{fig:vmf}.

Formally, let $f_\text{norm} = \|f\|$ and $f_\text{dir} = f / f_\text{norm}$.
Let $\vmf\left(v ; \mu, \kappa \right)$ denote a vMF distribution
over points $v$ on the unit sphere (i.e., directions) with mean vector $\mu$ 
and concentration parameter $\kappa$ (in such a distribution,
the log-likelihood of a point decays linearly with its cosine similarity to $\mu$,
and the rate of decay is controlled by $\kappa$).
Finally, define:
\begin{align*}
   q(z_\text{dir} \mid x',x) &= \vmf\left(z_\text{dir} ; f_\text{dir}, \kappa\right) \\
 q(z_\text{norm}\mid x',x) &= \text{Unif}(z_{\text{norm}} ; [\tilde{f}_{\text{norm}},\tilde{f}_{\text{norm}}+\epsilon])
\end{align*}
where $\tilde{f}_\text{norm} = \min(f_\text{norm}, 10 - \epsilon)$ is the truncated norm. 
The resulting edit vector is
$z = z_\text{dir} \cdot z_\text{norm}$.

The inverse neural editor $q$ is parameterized by the word vectors $\Phi$ and
has hyperparameters $\kappa$ and $\epsilon$.
Further details are provided in Section \ref{sec:q-discuss}.

\subsection{Details of the inverse neural editor}
\label{sec:q-discuss}

\newcommand{\qnab}{\nabla_{\Theta_q}}

\paragraph{Differentiating w.r.t. $\Theta_q$.}
To maximize our training objective, we must be able to compute
$\qnab \elbo = \qnab \Lgen - \qnab \LKL$.

To compute $\qnab \Lgen$, we use a reparameterization trick. Specifically,
we can rewrite $z \sim \qedit$ as $z = h(\alpha)$ where
$h$ is a deterministic function differentiable with respect to $\Theta_q$
and $\alpha \sim p(\alpha)$ is an auxiliary random variable not depending on $\Theta_q$
(the details of $h$ and $\alpha$ are given in Appendix \ref{sec:reparam}).
We can then write:
\begin{align*}
\qnab \Lgen &= \qnab \EE_{z \sim \qedit}\left[\log \pedit \right] \\
&= \EE_{\alpha \sim p(\alpha)}\left[ \qnab \log p_\text{edit}(x\mid x', h(\alpha)) \right].
\end{align*}
This moves the derivative inside the expectation.
The inner derivative can now be computed via standard backpropagation.

Next, we turn to $\qnab \LKL$. First, note that:
\begin{align}
\LKL &= \kl(q(z_\text{norm} | x', x) \| p(z_\text{norm})) \nonumber \\
 &+ \kl(q(z_\text{dir} | x', x) \| p(z_\text{dir})). \label{eq:kl}
\end{align}
It is easy to verify that the first KL term does not depend on $\Theta_q$. The second term has
the closed form
\begin{align}
  &\kl(\vmf(\mu, \kappa)\| \vmf(\mu, 0))  = \kappa  \frac{I_{d/2}(\kappa)+I_{d/2-1}(\kappa)\frac{d-2}{2\kappa}}{I_{d/2-1}(\kappa) - \frac{d-2}{2\kappa}} \nonumber \\
  &\quad - \log(I_{d/2-1}(\kappa)) - \log(\Gamma(d/2)) \nonumber\\
  &\quad + \log(\kappa)(d/2-1)-(d-2)\log(2)/2,
\label{eq:kl_vmf}
\end{align}
where $I_n(\kappa)$ is the modified Bessel function of the first kind, $\Gamma$ is the gamma function,
and $d$ is the dimensionality of $f$.
We can see that this too is constant with respect to $\Theta_q$ via the following intuition: both the KL divergence and the prior do not change under rotations, and thus we can see $\kl(\vmf(\mu, \kappa) \| \vmf(\mu,0))) = \kl(\vmf(e_1, \kappa) \| \vmf(e_1,0)))$ by rotating $\mu$ to the first canonical basis vector. Hence $\qnab \LKL = 0$.

\paragraph{Comparison with existing VAE encoders.}

Our design of $q$ differs from the typical choice of a
standard normal distribution \cite{bowman2016continuous,kingma2014variational} for two reasons:

First, by construction, edit vectors are sums of word vectors and since cosine
distances are traditionally used to measure distances between word vectors, it
would be natural to encode distances between edit vectors by the cosine
distance. The von-Mises Fisher distribution captures this idea, as the log
likelihood decays with cosine similarity.

Second, our design of $q$ allows us to explicitly control the tradeoff between
the two terms in our objective, $\Lgen$ and $\LKL$. Note from equations \ref{eq:kl} and
\ref{eq:kl_vmf} that $\LKL$ is purely a function of the hyperparameters $\epsilon$ and
$\kappa$, and can thus be controlled exactly. By taking $\kappa \to 0$ and
$\epsilon$ to the maximum norm, we can drive $\LKL$ arbitrarily close to 0. As a tradeoff,
smaller values of $\kappa$ produce a noisier edit vector, leading to
a smaller $\Lgen$. We find a good balance by tuning $\kappa$.

In contrast, when using a Gaussian variational encoder, the KL term takes a
different value per example and cannot be explicitly controlled. Consequently,
\newcite{bowman2016continuous} and others have observed that training tends to aggressively drive these KL terms to zero,
leading to uninformative values of $z$ --- even when
multiplying $\LKL$ by a carefully tuned and annealed importance weight.
 \section{Experiments}
\label{sec:experiments}

We divide our experimental results into two parts. In \refsec{gen-results}, we
evaluate the merits of the prototype-then-edit model as a generative modeling strategy,
measuring its improvements on language modeling (perplexity) and generation
quality (human evaluations of diversity and plausibility).
In \refsec{edit-results}, we focus on the semantics learned by the model and
 its latent edit vector space. We demonstrate that it possesses
 interpretable semantics, enabling us to smoothly control the magnitude of edits,
 incrementally optimize sentences for target properties, and
perform analogy-style sentence transformations.

\subsection{Datasets}
We evaluate perplexity on the Yelp review corpus 
(\textsc{Yelp}, Yelp \shortcite{yelp2017yelp}) and the One Billion Word
Language Model Benchmark (\textsc{BillionWord}, Chelba \shortcite{chelba2013one}). 
For qualitative evaluations of generation quality and semantics,
we focus on \textsc{Yelp} as our primary test case, as we found
that human judgments of semantic similarity were much better calibrated in this
focused setting.

For both corpora, we used the named-entity recognizer (NER) in spaCy\footnote{\url{honnibal.github.io/spaCy}} to replace named entities with their NER categories.
We replaced tokens outside the top 10,000
most frequent tokens with an ``out-of-vocabulary'' token.

\subsection{Generative modeling} \label{sec:gen-results}
\begin{figure*}[ht!]
    \vspace{-10pt}
  \centering
  \includegraphics[scale=0.21]{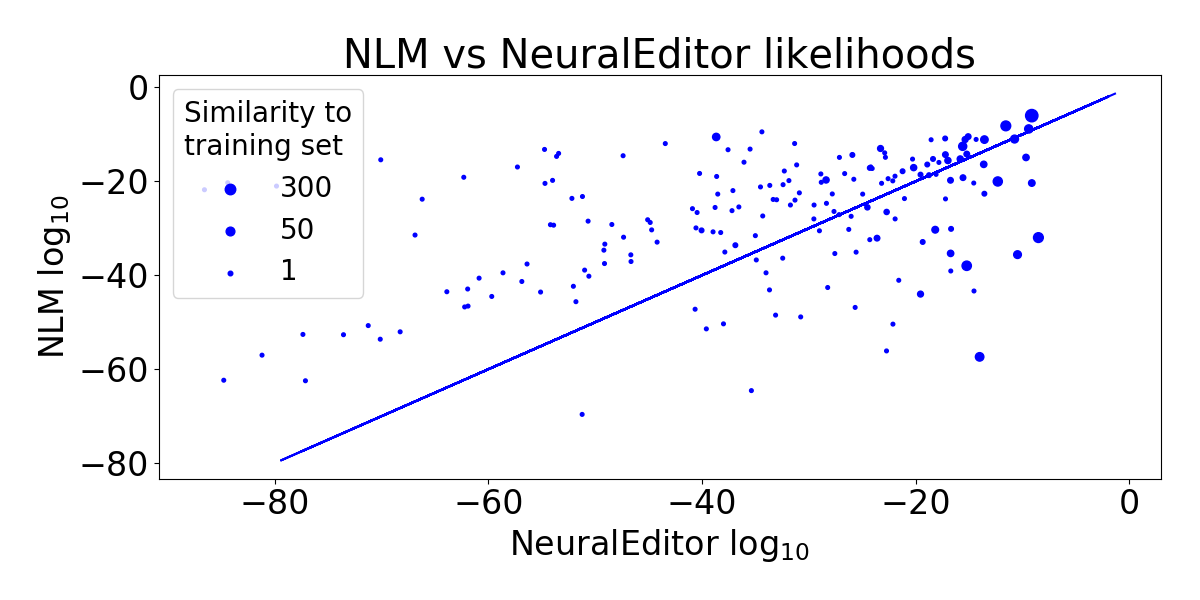}
  \includegraphics[scale=0.21]{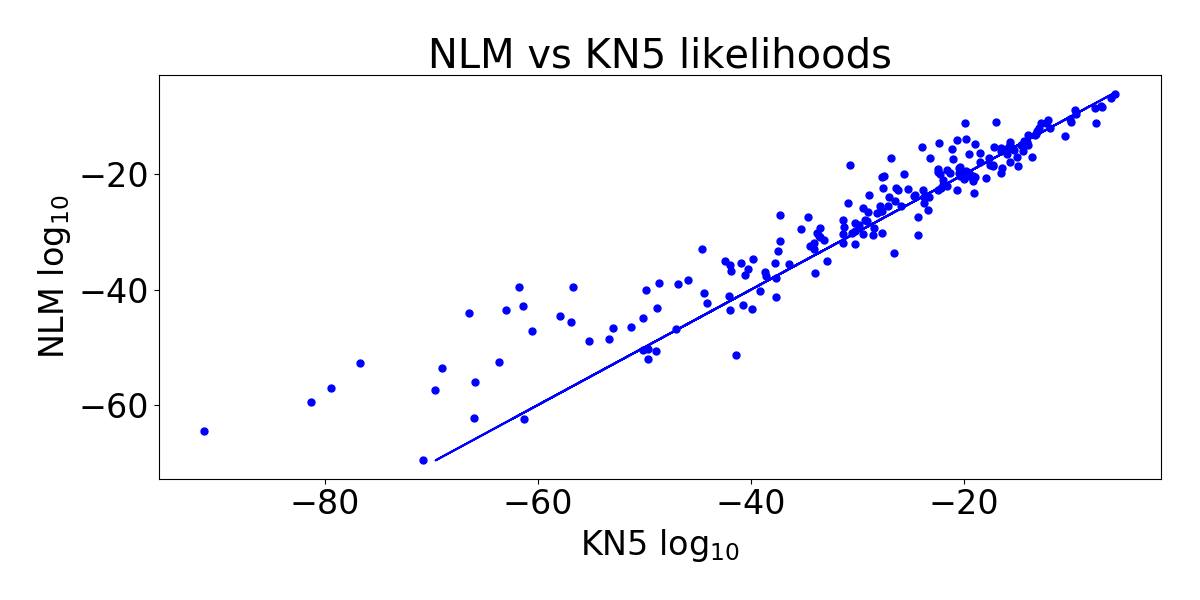}
  \vspace{-10pt}
  \caption{\textsc{NeuralEditor} outperforms \textsc{NLM} on examples similar to those in the training set (left panel, point size indicates number of training set examples with Jaccard distance < 0.5). The N-gram baseline (right) shows no such behavior, with \textsc{NLM} outperforming \textsc{KN5} on most examples.
}
  \vspace{-10pt}
  \label{fig:ngmix}
\end{figure*}

We compare \textsc{NeuralEditor} as a language model against the following baseline language models:
\begin{enumerate}
\itemsep0em
\item \textsc{NLM}: a standard left-to-right neural language model generating from scratch. For fair comparison, we use the exact same
architecture as the decoder of \textsc{NeuralEditor}.
\item \textsc{KN5}: a standard 5-gram Kneser-Ney language model in KenLM \cite{heafield2013scalable}.
\item \textsc{Memorization}: generates by sampling a sentence
  from the training set.
\end{enumerate}

\paragraph{Perplexity.} We start by evaluating \textsc{NeuralEditor}'s value as a language
model, measured in terms of perplexity. We use the likelihood lower bound in Equation \ref{eq:lowerbound}, where we sum over training set instances within Jaccard distance < 0.5, and for the VAE term in \textsc{NeuralEditor}, we use the one-sample approximation to the lower bound used in Kingma \shortcite{kingma2014variational} and Bowman \shortcite{bowman2016continuous}.

To evaluate \textsc{NeuralEditor}'s perplexity, we use linear smoothing with \textsc{NLM} to account for rare sentences not within our Jaccard distance threshold. This smoothing corresponds to occasionally sampling a special prototype sentence that can be edited into any other sentence and we use a smoothing weight of 0.1 (for full details, see Appendix \ref{sec:smoothing}). We find \textsc{NeuralEditor} improves perplexity over \textsc{NLM} and \textsc{KN5}. Table \ref{tab:perplex} shows that this is the case for both \textsc{Yelp} and the more general \textsc{BillionWord}, which contains substantially fewer test-set
sentences close to the training set. On \textsc{Yelp}, we surpass even the best
ensemble of \textsc{NLM} and \textsc{KN5}, while on \textsc{BillionWord} we nearly
match their performance.

Comparing each model at a per-sentence level, we see that \textsc{NeuralEditor} drastically improves log-likelihood for a significant number of sentences in
the test set (Figure \ref{fig:ngmix}). 
Proximity to a prototype seems to be the chief determiner of \textsc{NeuralEditor}'s performance.
\begin{table}[h]
\centering
\resizebox{0.42\textwidth}{!}{\begin{tabular}{l|C{1.5cm}|C{1.6cm}}
  Model &  Perplexity (Yelp) & Perplexity (\textsc{BillionWord})  \\
  \hline
  \textsc{KN5}  &   56.546 & 78.361\\
  \textsc{KN5}+\textsc{Memorization} & 55.184 & 73.468\\
  \hline
  \textsc{NLM}  & 39.026 & 55.146\\
  \textsc{NLM}+\textsc{Memorization} &  38.086 & 50.969\\
  \textsc{NLM}+\textsc{KN5} & 37.312 & \textbf{47.472}\\
  \textsc{NeuralEditor}($\kappa=0$) & \textbf{26.87} & 48.755\\
  \textsc{NeuralEditor}($\kappa=25$) & 27.41 & 48.921 \\
\end{tabular}}
\caption{Perplexity of \textsc{NeuralEditor} with the two VAE parameters $\kappa$ outperform all methods on \textsc{Yelp} and all non-ensemble methods on \textsc{BillionWord}.}
\label{tab:perplex}
\end{table}

Since \textsc{NeuralEditor} draws its strength from sentences in the training
set, we also compared against a simpler alternative, in which we ensemble
 \textsc{NLM} and \textsc{Memorization} (retrieval without edits). \textsc{NeuralEditor} performs
dramatically better than this alternative.
Table~\ref{tab:genedit} also qualitatively demonstrates that sentences generated
by \textsc{NeuralEditor} are substantially different from the original
prototypes.

\begin{table}
  \centering
  \resizebox{0.49\textwidth}{!}{
\begin{tabular}{p{50mm}| p{50mm} }
\textbf{Prototype} $x'$ &   \textbf{Revision} $x$ \\
  \hline
    this place gets <cardinal> stars for its diversity in its menu . & this place gets <cardinal> stars although not for the prices .\\
  \hline
great food and the happy hour deals were out of this world . & the deals are great and the food is out of this world .\\
    \hline
  i've been going to <person> for <date> and i used to really like this place . & i've been going to this place for <date> now and love it . \\
  \hline
   their food is great , and you can't beat the price . &  you can't beat the service and food here .\\
  \hline
\end{tabular}}
\caption{Edited generations are substantially different from the sampled prototypes.}
\label{tab:genedit}
\vspace{-5pt}
\end{table}

\paragraph{Human evaluation.}

We now turn to human evaluation of generation quality, focusing on
grammaticality and plausibility. We evaluated plausibility by asking human raters, ``How
plausible is it for this sentence to appear in the corpus?'' on a scale of 1--
3. We evaluate generations from \textsc{NeuralEditor} against an NLM with a 
temperature parameter on the per-token softmax\footnote{
If $s_i$ is the softmax logit for token $w_i$ and $\tau$ is a temperature parameter,
the temperature-adjusted distribution is $p(w_i) \propto \exp(s_i / \tau)$.}
as well as a baseline which generates sentences by randomly sampling from the training set and replacing synonyms, where the probability of substitution follows $\exp(s_{ij} / \tau)$, where $s_{ij}$ is the cosine similarity between the original word and its synonym according to GloVe word vectors. 

Decreasing the temperature parameter below 1 is a popular technique
for suppressing incoherent and ungrammatical sentences. Many NLM systems have noted an undesirable tradeoff between grammaticality and diversity, where a
temperature low enough to enforce grammaticality results in short and generic
utterances \cite{li2016diversity}.

\begin{figure*}
  \includegraphics[scale=0.26]{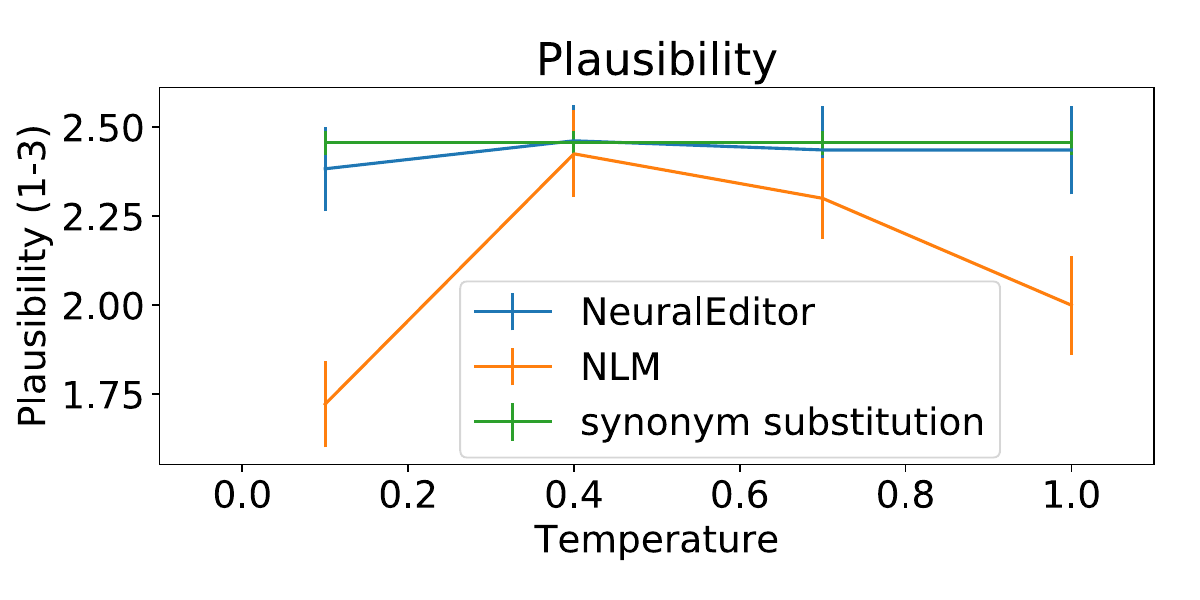}
  \includegraphics[scale=0.26]{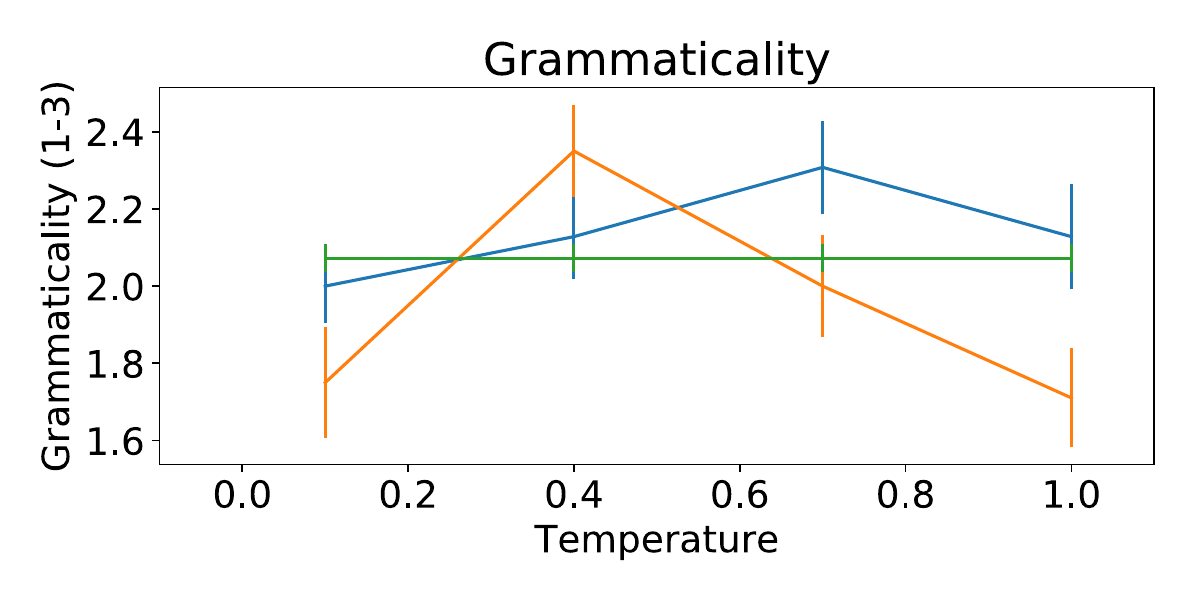}
  \includegraphics[scale=0.26]{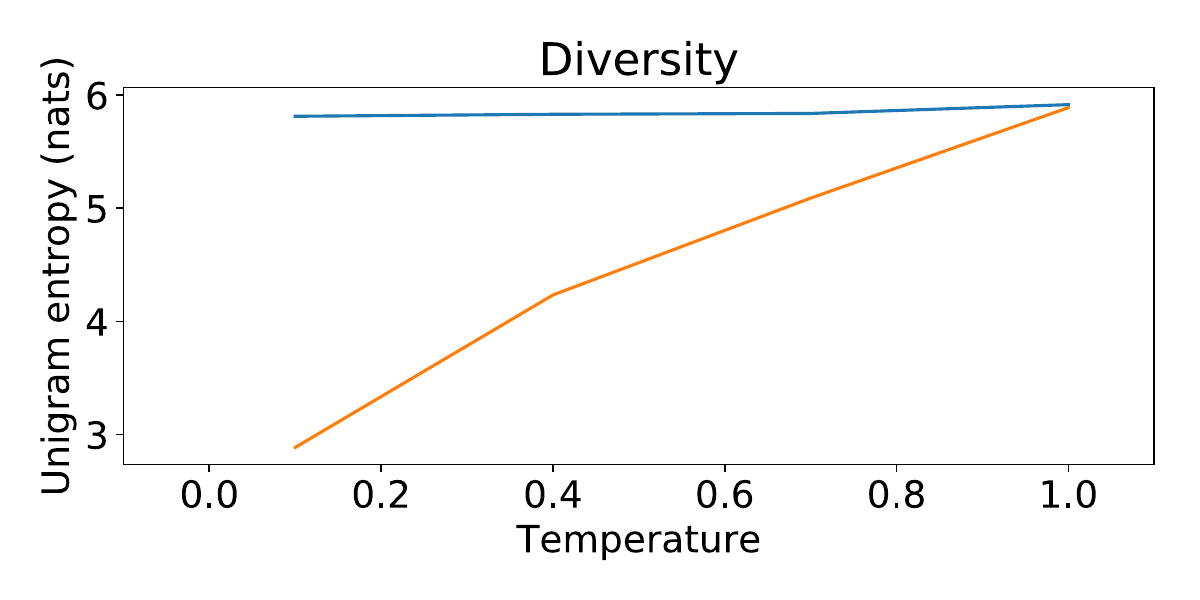}
  \vspace{-10pt}
  \caption{\textsc{NeuralEditor} provides plausibility and grammaticality on par with the
  best, temperature-tuned language model without any loss of diversity as a
  function of temperature. Results are based on 400 human evaluations.}
  \label{fig:qual_eval_turk}
\end{figure*}

Figure \ref{fig:qual_eval_turk} illustrates that both the grammaticality and
plausibility of \textsc{NeuralEditor} without any temperature annealing is on par with the best tuned temperature for \textsc{NLM}, with a far higher diversity, as measured by the discrete entropy over unigram frequencies. We also find that decreasing the temperature of \textsc{NeuralEditor} can be used to slightly improve the grammaticality, without substantially reducing the diversity of the generations.

Comparing with the synonym substitution model, we find both models have high plausibility, since synonym substitution maintains most of the words, but low grammaticality compared to both \textsc{NeuralEditor} and the NLM. Additionally, applying synonym substitutions to training examples has extremely low coverage -- none of the sentences in the test set can be generated via synonym substitution, and thus this baseline has higher perplexity than all other baselines in Table \ref{tab:perplex}.

A key advantage of edit-based models thus emerges: Prototypes sampled from the
training set organically inject diversity into the generation process, even if
the temperature of the decoder in \textsc{NeuralEditor} is zero. Hence, we can keep the decoder at a very low temperature to maximize grammaticality and plausibility, without sacrificing diversity. In contrast, a zero temperature \textsc{NLM} would collapse to outputting one generic sentence.

This also suggests that the temperature parameter for \textsc{NeuralEditor} captures a more natural notion of diversity --- a temperature of 1.0 encourages more aggressive
extrapolation from the training set while lower temperatures favor more
conservative mimicking. This is likely to be more useful than the tradeoff for generation-from-scratch, where low temperature also affects the diversity of generations.

\paragraph{Categorizing edits.}

To better understand the behavior of \textsc{NeuralEditor}, we measured the frequency with which random edits from \textsc{NeuralEditor} matched known syntactic transformations. 

We use the rule-based transformations defined in He \shortcite{he2015syntax} as our set of transformations to test, and search the corpus for sentences where these rules can be applied. We then apply the rule-based transformation, and measure the log-likelihood that \textsc{NeuralEditor} generates the transformed outputs. We find that the edit model assigns relatively high probability to the identity map (no edits), followed by simple reordering transformations such as reordering \emph{to/that} Clauses (\emph{It is }\texttt{ADJP}\emph{ to/that }\texttt{SBAR/S} $\to$ \emph{To }\texttt{S/BARS}\emph{ is }\texttt{ADJP}). Of the rules, active / passive receives the lowest probability, partially due to the rarity of passive voice sentences in the Yelp corpus (\reftab{edittype}). 

In all cases, the model assigns substantially higher probability to these rule-based transformations over editing to random sentences or shuffling the tokens randomly to match the Levenstein distance of each rule-based transform. 
\begin{table*}
  \centering
  \resizebox{0.95\textwidth}{!}{
    \begin{tabular}{p{45mm}| p{45mm} | p{55mm} | p{55mm}}
      \toprule
Type of edit & $-\log(p)$ per token & Example & Transformed example\\
               \hline
      Identity  & $0.33 \pm 0.006$ & It is important to remain watchful. & It is important to remain watchful.\\
        \specialrule{0.1pt}{0.1pt}{0.1pt}
  \emph{to} Clause reordering & $1.62 \pm 0.156$ & It is important to remain watchful.
                                              & To remain watchful is important.\\
              \specialrule{0.1pt}{0.1pt}{0.1pt}
  Quotative verb reordering & $2.02 \pm 0.0359$ & They announced that the president will restructure
the division. & The president will restructure the division,
                they announced.\\
                    \specialrule{0.1pt}{0.1pt}{0.1pt}
  Conjunction reversal & $2.52 \pm 0.0520$ & We should march because winter
is coming. & Winter is coming, because of
             this, we should march.\\
                          \specialrule{0.1pt}{0.1pt}{0.1pt}
      Genitive reordering & $2.678 \pm 0.0477$ & The best restaurant of New York.& New York's best restaurant.\\
                          \specialrule{0.1pt}{0.1pt}{0.1pt}
  Active / passive & $3.271 \pm 0.0298$ & The talk was denied by the boycott group
spokesman. & The boycott group spokesman denied the talk.\\
      \hline
      Random sentence reordering & $4.42 \pm 0.026$ & It is important to remain watchful. & It remain is to important watchful.\\
      Editing to random sentences & $6.068 \pm 0.084$ & &\\

\end{tabular}}
\caption{\textsc{NeuralEditor} assigns high probabilities to the syntactic transformations defined in He  \shortcite{he2015syntax} compared baselines of editing to random sentences or randomly reordering tokens to match the Levenstein distance of a syntactic edit. Small transformations, such as clause reordering, receive higher probability than more structural edits such as changing from active to passive voice.
  }
  \label{tab:edittype}
  \end{table*}

\subsection{Semantics of \textsc{NeuralEditor}} \label{sec:edit-results}

In this section, we investigate the learned semantics of \textsc{NeuralEditor}, focusing
on the two desiderata discussed in Section \ref{sec:problem}: semantic smoothness, and consistent edit behavior.

In order to establish a baseline for these properties, we consider existing sentence generation techniques
which can sample semantically similar sentences. The most similar language modeling approach which can capture semantics is the sentence variational autoencoder (\textsc{SVAE}) which imposes semantic structure onto a latent vector space, but uses the latent vector to represent the entire sentence, rather than just an edit.

To use the SVAE to ``edit'' a target sentence into a semantically similar sentence, we perturb its underlying latent sentence vector and then decode the
result back into a sentence --- the same method used in \newcite{bowman2016continuous}.

\paragraph{Semantic smoothness.}

A good editing system should have fine-grained control over the semantics of a
sentence: i.e., each edit should only alter the semantics of a sentence by a
small and well-controlled amount. We call this property semantic smoothness.

To study smoothness, we first generate an ``edit sequence'' by randomly
selecting a prototype sentence, and then repeatedly editing via \textsc{NeuralEditor}
(with edits drawn from the edit prior $p(z)$) to produce a sequence of
revisions. We then ask human annotators to rate the size of the semantic changes
between revisions. An example is given in Table \ref{tab:exwalk}.

\begin{table*}[h!]
   \centering
   \resizebox{\textwidth}{!}{
    \begin{tabular}{ll}
            \toprule
       \textsc{NeuralEditor} & \textsc{SVAE}\\
\hline
    this food was amazing one of the best i've tried, service was fast and great. &  this food was amazing one of the best i've tried, service was fast and great. \\
                   this is the best food and the best service i've tried in <gpe>. &                      this place is a great place to go if you want a quick bite. \\
             some of the best <norp> food i've had in <date> i've lived in <gpe>. &                                the food was good, but the service was terrible. \\
                     i have to say this is the best <norp> food i've had in <gpe>. &                                           this is the best <norp> food in <gpe>. \\
                           best <norp> food i've had since moving to <gpe> <date>. &                            this place is a great place to go if you want to eat. \\
                      this was some of the best <norp> food i've had in the <gpe>. &                                           this is the best <norp> food in <gpe>. \\
\bottomrule
\end{tabular}
}
\caption{Example random walks from \textsc{NeuralEditor}, where the top sentence is the prototype.}
\label{tab:exwalk}
\end{table*}

We compare to two baselines, one based upon the sentence variational autoencoder (\textsc{SVAE}) and another baseline which simply samples similar sentences from the training set according to average word vector similarity (\textsc{Cosine}).

For \textsc{SVAE}, we generate a similar sequence of sentences by first encoding the prototype sentence, and then decoding after the addition of a random Gaussian with variance 0.4.\footnote{The variance was selected so that \textsc{SVAE} and \textsc{NeuralEditor} have the same average human similarity judgement between two successive sentences. This avoids situations where \textsc{SVAE} produces completely unrelated sentence due to the perturbation size.} This process is repeated to produce a sequence of sentences which we can view as the \text{SVAE} equivalent of the edit sequence.

For \textsc{Cosine}, we generate sentences from the training set using exponentiated cosine similarity between averaged word vectors. The temperature parameter for the exponential was selected as before to match the average human similarity judgement. 

\begin{figure}[h!]
      \centering
  \includegraphics[scale=0.24,trim={0 15 0 0}, clip]{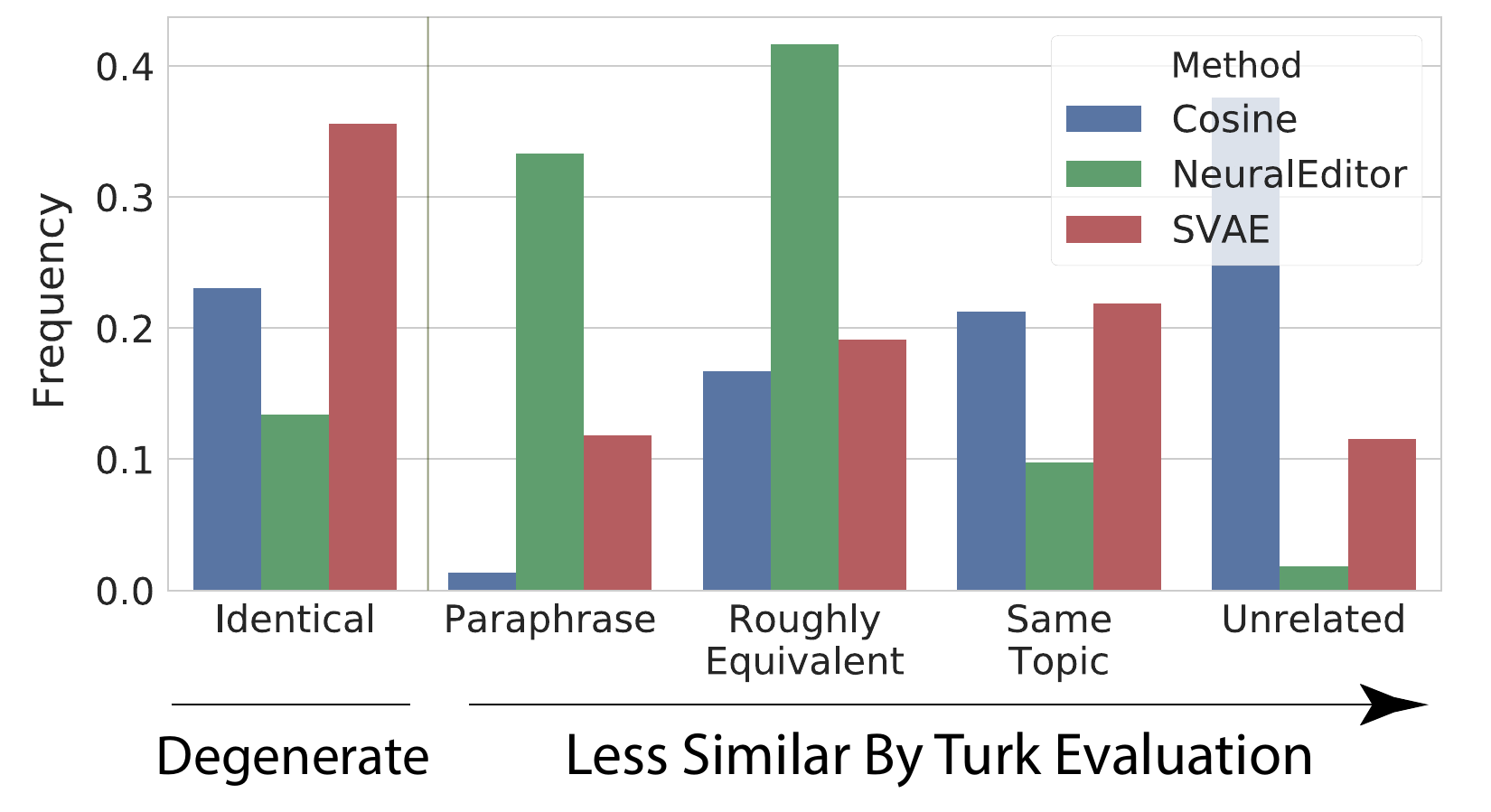}
  \caption[test]{Compared with baselines, \textsc{NeuralEditor} frequently generates paraphrases and similar sentences while avoiding unrelated and degenerate ones.\protect\footnotemark}
  \label{fig:rwalk}
  \end{figure}

Figure \ref{fig:rwalk} shows that \textsc{NeuralEditor} frequently generates paraphrases despite being trained on lexical similarity, and only 1\% of edits are unrelated from the prototype. In contrast,  \textsc{SVAE} often repeats sentences exactly, and when it makes an edit it is equally likely to generate unrelated sentences. \textsc{Cosine} performs even worse likely due to the difficulty of retrieving similar sentences for rare and long sentences. 

Qualitatively (Table \ref{tab:exwalk}), \textsc{NeuralEditor} seems to generate long, diverse
sentences which smoothly change over time, while the SVAE biases towards
short sentences with several semantic jumps, presumably due to the difficulty of training a sufficiently informative SVAE encoder.

  \footnotetext{
    545 similarity assessments pairs were collected
    through Amazon Mechanical Turk following
    Agirre \shortcite{agirre2014semeval}, with the same scale and prompt. Similarity judgements were converted to
    descriptions by defining Paraphrase (5), Roughly Equivalent (4-3), Same Topic (2-1), Unrelated (0).
  }

  \begin{figure*}[h!]
      \vspace{-10pt}
  \centering
  \includegraphics[width=0.3\textwidth]{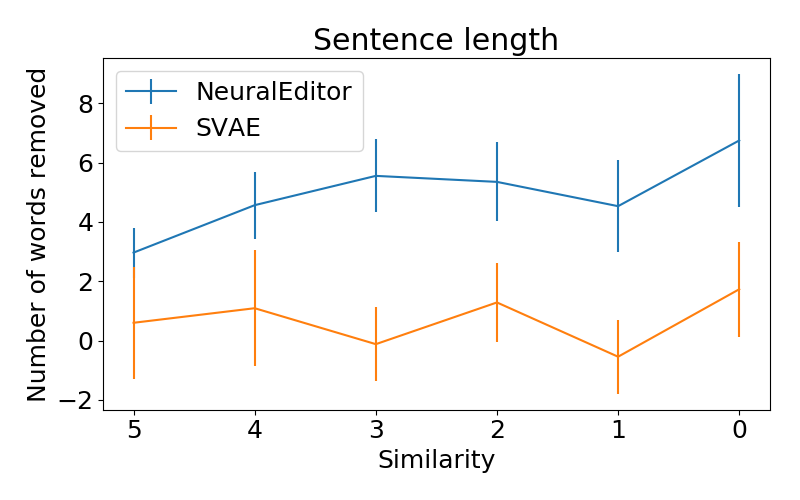}
  \includegraphics[width=0.3\textwidth]{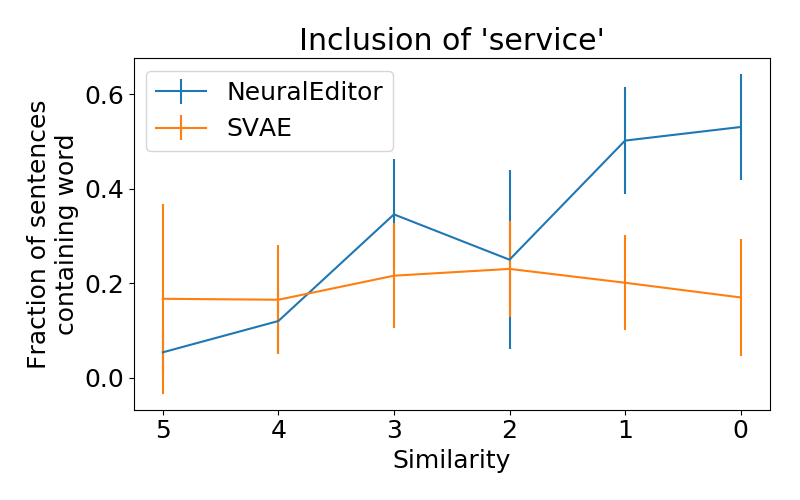}
  \includegraphics[width=0.3\textwidth]{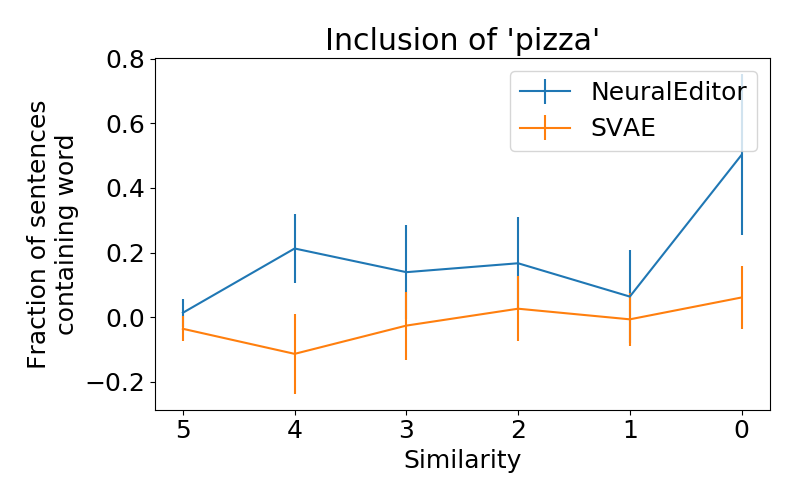}
    \vspace{-10pt}
  \caption{\textsc{NeuralEditor} can shorten sentences (left), include common words (center, the word `service') and rarer words (right `pizza') while maintaining similarity.}
  \label{fig:opt}

\end{figure*}

\begin{table*}[h!]
  \centering
    \resizebox{0.95\textwidth}{!}{
  \begin{tabular}{ll}
    \toprule
    \textsc{NeuralEditor} & \textsc{SVAE}\\
\midrule
 the coffee ice cream was one of the best i've ever tried. &  the coffee ice cream was one of the best i've ever tried. \\
                some of the best ice cream we've ever had! &              the <unk> was very good and the food was good. \\
              just had the best ice - cream i've ever had! &                          the food was good, but not great. \\
                  some of the best \textbf{pizza} i've ever tasted! &                          the food was good, but not great. \\
     that was some of the best \textbf{pizza} i've had in the area. &            the food was good, but the service was n't bad. \\
\bottomrule
  \end{tabular}
}
  \caption{Examples of word inclusion trajectories for `pizza'. \textsc{NeuralEditor} produces smooth chains that lead to word inclusion, but the SVAE gets stuck on generic sentences.}
\label{tab:control}
\end{table*}

\paragraph{Smoothly controlling sentences.}
We now show that we can selectively choose edits sampled from \textsc{NeuralEditor} to
incrementally optimize a sentence towards desired attributes.
This task serves as a useful measure of semantic
coverage: if an edit model has high coverage over sentences that are semantically similar
to a prototype, it should be able to satisfy the target attribute while deviating minimally
from the prototype's original meaning.

We focus on controlling two simple attributes: compressing a sentence to below a
desired length (e.g., 7 words), and inserting a target keyword into the sentence
(e.g., ``service'' or ``pizza'').

Given a prototype sentence, we try to discover a semantically similar sentence
satisfying the target attribute using the following procedure:
First, we generate 1,000 edit sequences using the procedure described earlier.
Then, we select the sequence with highest likelihood whose endpoint possesses the target attribute.
We repeat this process for a large number of prototypes.

We use almost the same procedure for the SVAE, but instead of selecting by
highest likelihood, we select the sequence whose endpoint has shortest latent
vector distance from the prototype (as this is the SVAE's metric of semantic
similarity).

In Figure \ref{fig:opt}, we then aggregate the sentences from the collected edit sequences,
and plot their \emph{semantic similarity to the prototype}
against their \emph{success in satisfying the target attribute}.
Not surprisingly, as target attribute satisfaction rises, semantic similarity drops.
However, we also see that \textsc{NeuralEditor} sacrifices less semantic similarity 
to achieve the same level of attribute satisfaction as \textsc{SVAE}.
\textsc{SVAE} is reasonable on tasks
involving common words (such as the word \textit{service}), but fails when the model is asked to generate rarer words such
as \textit{pizza}. Examples from these word inclusion problems show
that \textsc{SVAE} often becomes stuck generating short, generic sentences (Table \ref{tab:control}).

\begin{table*}[ht!]
  \vspace{-10pt}
\resizebox{\textwidth}{!}{
\begin{tabular}{rrrrrrrrrl}
  \toprule
  \multicolumn{3}{c}{Google} & \multicolumn{6}{c}{Microsoft} & Method\\
 \cmidrule(lr){1-3}  \cmidrule(lr){4-9} 
  gram4-superlative &   gram3-comparative &  family & JJR\_JJS &     VB\_VBD &  VBD\_VBZ &  NN\_NNS &  VB\_VBZ &  JJ\_JJR &        \\
  \midrule
  0.45 & 0.85 & 1.0 & 0.75 & 0.63 & 0.82 & 0.82 & 0.61 & 0.77 & GloVE\\
  \midrule
  0.75 & 0.75 & 0.29 & 0.79 & 0.57 & 0.60 & 0.58 & 0.41 & 0.24 &  Edit vector (top 10)\\
    0.60 & 0.32 & 0.01 & 0.45 & 0.16 & 0.23 & 0.17 & 0.01 & 0.06 & Edit vector (top 1)\\
    0.10 & 0.10 & 0.09 & 0.10 & 0.08 & 0.14 & 0.15 & 0.05 & 0.03 & Sampling (top 10)\\
\bottomrule
\end{tabular}
}
\caption{Edit vectors capture one-word sentence analogies with performance close to lexical analogies.
  }
\label{tab:anal}
\end{table*}

\begin{table*}[h!]
    \centering
  \resizebox{0.7 \textwidth}{!}{
    \begin{tabular}{ r  | c  c  }
      & Example 1 & Example 2 \\ 
      \hline
Context   & he comes home tired and happy . & i went with a larger group to <person> 's .\\
Edit  &  + was - is    &  + good - better\\ 
Result   &  = he came home happy and tired .
                                  &  = i went to <person> 's with a large group .\\
\end{tabular}}
\caption{Examples of lexical analogies correctly answered by \textsc{NeuralEditor}. Sentence pairs generating the analogy relationship are shortened to only their lexical differences.}
\label{tab:analcorrect}
\vspace{-10pt}
\end{table*}

\paragraph{Consistent edit behavior: sentence analogies.}

In the previous results, we showed that edit models learn to generate
semantically similar sentences. We now assess whether the edit vector possesses
globally consistent semantics. Specifically, applying the same edit vector to
different sentences should result in semantically analogous edits.

For example, if we have an edit vector which edits the sentence $x_1=\phantom{}$``this was a good
restaurant'' into $x_2=\phantom{}$``this was the best restaurant''. Given a new sentence $y_1=\phantom{}$``The cake was great'',
we expect applying the same edit vector to result in  $y_2=\phantom{}$``The cake was the greatest''.

Formally, suppose we have two sentences, $x_1$ and $x_2$, which are related by
some underlying semantic relation $r$. Given a new sentence $y_1$, we would like
to find a $y_2$ such that the same relation $r$ holds between $y_1$ and $y_2$.

Our approach is to estimate the edit vector between $x_1$ and $x_2$ as
$\hat{z} = f(x_1, x_2)$ --- the mode of the inverse neural editor $q$.
We then apply this edit vector to $y_1$ using the neural editor to yield $\hat{y_2} = \text{argmax}_x p_\text{edit}(x \mid y_1, \hat{z})$.

Since it is difficult to output $\hat{y_2}$ exactly matching $y_2$, we take the top
$k$ candidate outputs of $p_\text{edit}$ (using beam search) and evaluate whether
the gold $y_2$ appears among the top $k$ elements.

We generate the semantic relations $r$ using prior evaluations for
 word analogies \cite{mikolov2013efficient,mikolov2013linguistic}.
We leverage these to generate a new dataset of sentence analogies, using a simple strategy:
given an analogous word pair $(w_1, w_2)$, we mine the Yelp corpus for sentence pairs $(x_1, x_2)$
such that $x_1$ is transformed into $x_2$ by inserting $w_1$ and removing $w_2$ (allowing
for reordering and inclusion/exclusion of stop words).

For this task, we initially compared against the SVAE, but it had a top-$k$ accuracy close
to zero. Hence, we instead compare to \textsc{Sampling} which is a baseline which randomly samples an edit vector $\hat{z} \sim p(z)$,
instead using $\hat{z}$ derived from $f(x_1,x_2)$.

We also compare our accuracies to the simpler task of solving the word, rather than sentence-level analogies in \cite{mikolov2013efficient} using GloVe.
This task is substantially simpler, since the goal is to identify a single word (such as ``good:better::bad:?'') instead of an entire sentence.
Despite this, the top-10 performance of our model in Table~\ref{tab:anal} is
nearly as good as the performance of GloVe vectors on the simpler lexical
analogy task. In some categories, \textsc{NeuralEditor} at top-10 actually performs better than word
vectors, since \textsc{NeuralEditor} has an understanding of which words are likely
to appear in the context of a Yelp review. Examples in Table
\ref{tab:analcorrect} show the model is accurate and captures lexical analogies
requiring word reorderings.

 \section{Related work and discussion}

Our work connects with a broad literature on attention-based neural models,
retrieval-augmented text generation,
semantically meaningful representations, and nonparametric statistics.

Based upon recurrent neural
networks and sequence-to-sequence architectures~\cite{sutskever2014sequence},
neural language models~\cite{bengio2003neural} 
have been widely used due to their flexibility and performance across a wide
range of NLP tasks \cite{kalchbrenner2013recurrent,hahn2000challenges,ritter2011data}.
Our work is motivated by an emerging consensus
that attention-based mechanisms \cite{bahdanau2015neural} can substantially
improve performance on various sequence to sequence tasks by capturing more
information from the input sequence \cite{vaswani2017attention}.
Our work extends the applicability of attention mechanisms beyond
sequence-to-sequence models by allowing models to attend to randomly sampled sentences.

There is a growing literature on applying retrieval mechanisms to augment text generation models.
For example, in the image captioning literature, Hodosh \shortcite{hodosh2013framing},
Kuznetsova \shortcite{kuznetsova2013generalizing} and Mason \shortcite{mason2014domain}
proposed to generate image captions by first retrieving a prototype caption based on an image context, and then applying sentence compression to tailor the prototype to a particular
image. More recently, Song \shortcite{song2016retrieval} ensembled a retrieval
system and an NLM for dialogue, using the NLM to transform the retrieved
utterance, and Gu \shortcite{gu2017search} used an off-the-shelf search engine
system to retrieve and condition on training set examples. Although these approaches also edit text from the training set, these papers solve a fundamentally different problem since they solve conditional generation problems, and retrieve prototypes based on a context, where as our task is unconditional and thus there is no context which we can use to retrieve.

Our work treats the prototype $x'$ as a
latent variable rather than being given by a retrieval mechanism,
and marginalizes over all possible prototypes --- a challenge
which motivates our new lexical similarity training method in Section \ref{sec:lowerbound}.
Practically, marginalization over $x'$ makes our model attend to training examples based on similarity of \emph{output sequences},
while prior retrieval models attend to examples based on similarity of the \emph{input sequences}.

In terms of generation techniques that capture semantics, the sentence variational autoencoder (SVAE)
\cite{bowman2016continuous} is closest to our work in that it attempts
to impose semantic structure on a latent vector space. However, the SVAE's
latent vector is meant to represent the entire sentence, whereas the neural editor's
latent vector represents an edit. Our results from \refsec{edit-results} suggest
that local variation over edits is easier to model than global variation over sentences.

Our use of lexical similarity neighborhoods is
comparable to context windows in word vector training \cite{mikolov2013efficient}.
More generally, results in manifold learning demonstrate that a weak metric such as
lexical similarity can be used to extract semantic similarity through distributional
statistics \cite{tenenbaum2000global,hashimoto2016word}.

From a generative modeling perspective, editing randomly sampled training sentences closely resembles nonparametric kernel density estimation \cite{parzen1962} where one samples points from a training set, and adds noise to smooth the density. Our edit model is the text equivalent of Gaussian noise, and our training mechanism is a type of learned smoothing kernel.

Prototype-then-edit is a semi-parametric approach that remembers the entire
training set and uses a neural editor to generalize meaningfully beyond the training set.
The training set provides a strong inductive bias --- that the corpus
can be characterized by prototypes surrounded by semantically similar sentences
reachable by edits.  Beyond improvements on generation quality as measured by perplexity,
the approach also reveals new semantic structures via the edit vector.

\textbf{Reproducibility.} All code, data and experiments are available on the CodaLab platform at {\small \url{https://bit.ly/2rHsWAX}}.

\textbf{Acknowledgements.}
We thank the reviewers and editor for their insightful comments.
This work was funded by DARPA CwC program under ARO prime contract no. W911NF-15-1-0462.

\section{Appendix}
\paragraph{Construction of the LSH.}
\label{sec:lsh}
The LSH maps a sentence to lexically similar sentences in the corpus,
representing a graph over sentences.
We apply breadth-first search
(BFS) over the LSH sentence graph started at randomly selected seed sentences and uniformly
sample this set to form the training set. 
\paragraph{Reparameterization trick for $q$.}
\label{sec:reparam}
First, note that we can write $z_\text{norm} \sim q(z_\text{norm} | x', x)$ as
$z_\text{norm} = h_\text{norm}(\alpha_\text{norm}) \eqdef \tilde{f}_\text{norm} + \alpha_\text{norm}$
where $\alpha_\text{norm} \sim \text{Unif}(0, \epsilon)$.
Furthermore, \newcite{wood1994simulation} present a function $h_\text{dir}$
and auxiliary random variable $\alpha_\text{dir}$, such that $z_\text{dir} = h_\text{dir}(\alpha_\text{dir})$
is distributed according to a vMF with mean $f$ and concentration $\kappa$.
We can then define
$z = h(\alpha) \eqdef h_\text{dir}(\alpha_\text{dir}) \cdot h_\text{norm}(\alpha_\text{norm})$.

\paragraph{Smoothing for language models.}
\label{sec:smoothing}
\begin{figure}[h!]
    \vspace{-15pt}
  \centering
  \includegraphics[scale=0.20,trim={0 25 0 0},clip]{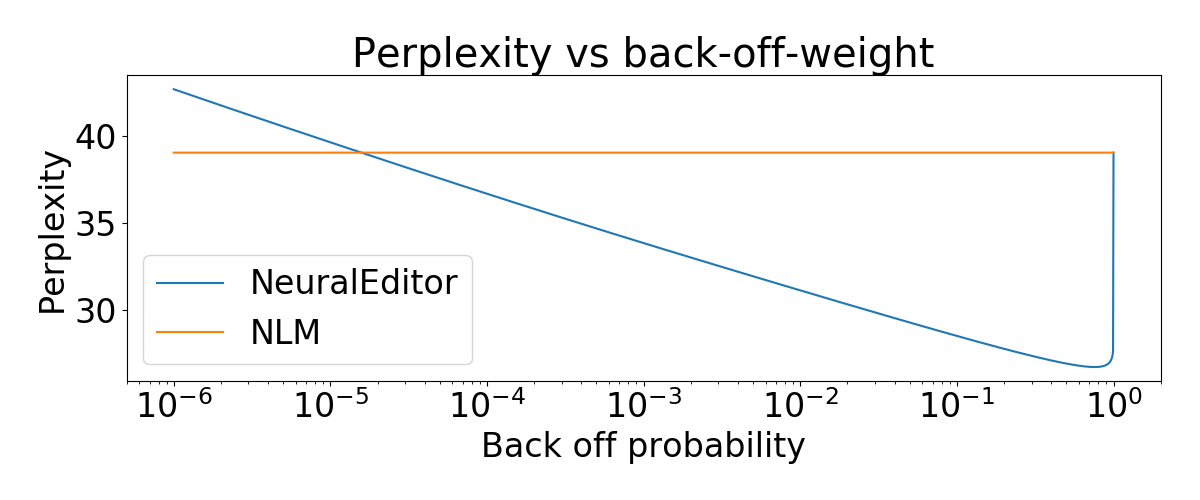}
  \caption[test]{Small amounts of smoothing are sufficient to make \textsc{NeuralEditor} outperform the baseline NLM.}
  \label{fig:lmsmooth}
\end{figure}
As a language model, \textsc{NeuralEditor} does not place probability on any test sentence which is sufficiently dissimilar from all training set sentences. In order to avoid this problem, we can consider a special prototype sentence `$\emptyset$' which can be edited into any sentence, and draw this special prototype with probability $p_\emptyset$. Concretely, we write:

\resizebox{0.47\textwidth}{!}{
\begin{minipage}{1.2\linewidth}
\vspace{-5pt}
\begin{align*}
  p(x) &=  \sum_{x'\in \mathcal{X} \cup \{\emptyset\}} p_{\text{edit}}(x|x')p_{\text{prior}}(x') \\
       &=  (1-p_\emptyset) \sum_{x'\in \mathcal{X}} \frac{1}{|\mathcal{X}|}p_{\text{edit}}(x|x') + p_\emptyset    \phantom{.}p_{\text{NLM}}(x).
\end{align*}
\vspace{2pt}
\end{minipage}}

This linearly smoothes between our edit model $(p_{\text{edit}})$ and the NLM $(p_{\text{NLM}})$ since our decoder is identical to the NLM, and thus conditioning on the special $\emptyset$ token reduces to using a NLM.
  
Empirically, we observe that even small values of $p_\emptyset$ produces low perplexity (Figure \ref{fig:lmsmooth}) corresponding to the observation that smoothing of \textsc{NeuralEditor} is only necessary to avoid degenerate log-likelihoods on a very small subset of the test set.

\bibliographystyle{acl2012}
\bibliography{refdb/all}

\end{document}